\documentclass[11pt]{article}
\usepackage[utf8]{inputenc}
\usepackage{microtype}
\usepackage[english]{babel}
\usepackage{framed}
\usepackage{fullpage}
\usepackage{hyperref}
\usepackage{xcolor}
\colorlet{darkblue}{blue!50!black}
\hypersetup{breaklinks=true,colorlinks=true,allcolors=darkblue}
\usepackage{heuristica}
\usepackage[heuristica,vvarbb,bigdelims]{newtxmath}
\usepackage{ClearSans}
\usepackage{inconsolata}
\usepackage[T1]{fontenc}
\usepackage{amsmath}
\usepackage{mathtools}
\usepackage{bm}
\usepackage{graphicx}
\usepackage{enumitem}

\usepackage{natbib}

\DeclareMathOperator{\conv}{conv}
\newcommand{\ZZ}{\mathcal{Z}}
\newcommand{\pfrac}[2]{\frac{\partial #1}{\partial #2}}

\title{Notes on Latent Structure Models and SPIGOT}
\author{André
F.~T.~Martins\\\href{mailto:andre.martins@unbabel.com}{\texttt{andre.martins@unbabel.com}}
\and Vlad Niculae\\\href{mailto:vlad@vene.ro}{\texttt{vlad@vene.ro}}}
\date{}
\begin{document}

\maketitle

\begin{abstract}
These notes aim to shed light on the recently proposed structured projected intermediate gradient optimization technique (SPIGOT, \citet{peng2018backpropagating}). SPIGOT is a variant of the straight-through estimator \citep{bengio_estimating_2013} which bypasses gradients of the argmax function by back-propagating a surrogate ``gradient.'' We provide a new interpretation to the proposed gradient and put this technique into perspective, linking it to other methods for training neural networks with discrete latent variables. As a by-product, we suggest alternate variants of SPIGOT which will be further explored in future work.
\end{abstract}

\section{Introduction}

In these notes, we assume a general latent structure model involving input variables $x \in \mathcal{X}$, output variables $y \in \mathcal{Y}$, and latent discrete variables $z \in \mathcal{Z}$.
We assume that $\mathcal{Z} \subseteq \{0,1\}^K$, where $K \le |\mathcal{Z}|$ (typically, $K \ll |\mathcal{Z}|$): i.e., the latent discrete variable $z$ can be represented as a $K$-th dimensional binary vector. This often results from a decomposition of a structure into parts: for example, $z$ could be a dependency tree for a sentence of $L$ words, represented as a vector of size $K=O(L^2)$, indexed by pairs of word indices $(i,j)$,
with $z_{ij} = 1$ if arc $i \rightarrow j$ belongs to the tree, and $0$ otherwise.

%As a simplifying assumption, we start by considering the case where $z$ is an \emph{unstructured} categorical variable,
%i.e., $\mathcal{Z} = \{e_1, \ldots, e_K\}$, where each $e_k$ denotes a one-hot vector representing a category, whose $k$-th coordinate is 1 and all the others are zero.
%i.e., $K=|\mathcal{Z}|$ is the number of possible categories (labels), and each $z \in \mathcal{Z}$ is a one-hot vector representing a category.

\paragraph{Notation.} In the following, we denote the $(|\mathcal{Z}|-1)$-dimensional probability simplex by $\Delta^{|\mathcal{Z}|} = \{p \in \mathbb{R}^{|\mathcal{Z}|} \mid p \ge 0, \sum_{z \in \mathcal{Z}} p(z) = 1\}.$
Given $p \in \Delta^{|\mathcal{Z}|}$, we denote the expectation of a function
$h:\mathcal{Z}\rightarrow \mathbb{R}^D$ under the probability distribution $p$
by $\mathbb{E}_{z \sim p}[h(z)] = \sum_{z \in \mathcal{Z}} p_z h(z)$.
We denote the convex hull of the (finite) set $\mathcal{Z} \subseteq \mathbb{R}^K$ by $\mathrm{conv}(\mathcal{Z}) = \left\{\mathbb{E}_{z \sim p}[z] \mid p \in \Delta^{|\mathcal{Z}|}\right\}$.
The set $\mathrm{conv}(\mathcal{Z})$ can be interpreted as the smallest convex set which contains $\mathcal{Z}$.
%Note that $\mathrm{conv}(\{e_1, \ldots, e_K\}) = \Delta^{|K|}$.

\paragraph{Background.}
In the literature on structured prediction, the set $\conv(\ZZ)$ is sometimes
called the \emph{marginal polytope}, since any point inside it can be
interpreted as some marginal distribution over parts of the structure (arcs)
under some distribution over structures.
There are three relevant problems that may be formulated in a structured
setting:
\begin{itemize}
\item Finding the highest scoring structure, a.k.a. maximum a-posteriori (MAP):
identify
\begin{equation}\arg\max_{\mu \in \conv(\ZZ)} s^\top\mu\end{equation}

\item Marginal inference: finding the (unique) marginal distribution induced by
the scores $s$, given by an entropy projection onto the marginal polytope:
\begin{equation}\arg\max_{\substack{p \in \Delta^{|\ZZ|} \\ \mu =
\mathbb{E}_p[z]}} s^\top\mu \underbrace{-\sum_{z \in \ZZ} p_z \log p_z}_{H(p)}
\end{equation}
\item SparseMAP: finding the (unique) marginal distribution induced by
the scores $s$, given by an \textbf{Euclidean} projection onto the marginal polytope:
\begin{equation}\arg\max_{\mu \in \conv(\ZZ)} s^\top\mu - \frac{1}{2} \|\mu\|^2
\end{equation}
\end{itemize}

%Finding the highest scoring structure is known as \emph{maximum a posteriori}
%(MAP) decoding and can be written equivalently as an indexed search or as a
%linear program:
%\[
%\max_{z \in \ZZ} s^\top z = \max_{\mu \in \conv(\ZZ)} s^\top\mu.
%\]
%When there are no ties, the solution sets ($\arg\max$) of the two problems are
%identical. The structured entropy of a marginal distribution $\mu$ is defined
%in terms of the Shannon entropy of the corresponding distribution over
%structures:
%\[
%\bar{H}(\mu) = \max_{p \in \Delta^{|\ZZ|}} \underbrace{-\sum_{z \in \ZZ} p(z)
%\log p(z)}_{H(p)} \quad
%\text{subject to} \quad \mathbb{E}_{z\sim p}[z] = \mu.
%\]
%The entropy-regularized linear problem
%\[
%\arg\max_{\mu \in \conv(\ZZ)} s^\top\mu + \bar{H}(\mu)
%\]
%is known as \emph{Marginal inference}, and provides a unique mapping between
%score vectors $s$ and marginal distributions $\mu$. Similarly

\paragraph{Unstructured setting.}
We may encode the simple case of an unstructured categorical case by setting
$\ZZ = \{ e_1, \dots, e_K\}$ which leads to $\conv(\ZZ) = \Delta^K.$
The optimization problems above then recover some well known transformations, as
described in the table below.
\begin{table}[h]
\centering
\begin{tabular}{r l l}
& unstructured & structured \\
\hline
vertices & $e_k$ & $z_k$ \\
interior points & $p$ & $\mu$ \\
maximization & argmax & MAP \\
expectation & softmax & Marg \\
euclidean projection & sparsemax & SparseMAP \\
\hline
\end{tabular}

\end{table}

\section{Latent structure model}

Throughout,
we assume a neural network classifier, parametrized by $\phi$ and $\theta$, which consists of three parts:
\begin{itemize}
    \item An {\bf encoder function} $f_{\phi}$ which, given an input  $x \in \mathcal{X}$, outputs a vector of ``scores'' $s \in \mathbb{R}^K$, as $s = f_{\phi}(x)$;
    \item An {\bf argmax node} which, given these scores, outputs the highest-scoring structure:
    \begin{equation}\label{eq:argmax}
    \hat{z}(s) = \arg\max_{z \in \mathcal{Z}} s^\top z.
    \end{equation}
    %\item A transformation $\rho : \mathbb{R}^K \rightarrow \Delta^K$ which converts these scores into probabilities, as $p = \rho(s)$. Common examples are $\rho = \mathrm{argmax}$, which returns a one-hot vector indicating the class with the highest probability, and $\rho = \mathrm{softmax}$, which returns the probability of each class according to a Gibbs distribution.
    %We will also consider the case where this transformation is stochastic, in which case $z$ is sampled from a distribution with label probabilities $\rho(s)$.
    \item A {\bf decoder function} $g_{\theta}$ which, given $x \in \mathcal{X}$ and $z \in \mathcal{Z}$, makes a prediction $\hat{y} \in \mathcal{Y}$ as $\hat{y} = g_{\theta}(x, z)$. We will sometimes write $\hat{y}(z)$ to emphasize the dependency on $z$.
    For reasons that will be clear in the sequel, we assume that the decoder also accepts \emph{convex combinations} of latent variables as input, i.e., it may also output predictions $g_{\theta}(x, \mu)$ where $\mu \in \mathrm{conv}(\mathcal{Z})$.
    %\item A decoder function $g_{\theta}$ which, given $x \in \mathcal{X}$ and $p \in \Delta^K$, makes a prediction $\hat{y} \in \mathcal{Y}$ as $\hat{y} = g_{\theta}(x, p)$. We will often write $\hat{y}(p)$ to emphasize the dependency on $p$.
\end{itemize}
Thus, given input $x \in \mathcal{X}$, this network predicts:
\begin{equation}
    \hat{y} = g_\theta\left(x, \overbrace{\arg\max_{z \in \mathcal{Z}} f_\phi(x)^\top z}^{\hat{z}(s)}\right).
\end{equation}
To train this network, we assume a loss function  $L(\hat{y}, y^\star)$, where $y^\star$ denotes the true output. We want to minimize this loss over the training data, using the gradient backpropagation algorithm.

%A challenge is that, in order to backpropagate gradients to the encoder parameters $\phi$, we need the gradient
%$\frac{\partial \rho(s)}{\partial s}$. However, in the deterministic case, if $\rho = \mathrm{argmax}$, this gradient is zero almost everywhere; and in the stochastic case, it is not even defined.

We assume $\nabla_\theta L(\hat{y}, y^\star)$ is easy to compute: it can be done by performing standard gradient backpropagation from the output layer until the output of the argmax node.
The main challenge of this model is to sidestep the argmax node to propagate gradient information to the encoder parameters. Indeed, we have:
\begin{equation}
    \nabla_\phi L(\hat{y}, y^\star) = \frac{\partial f_\phi(x)}{\partial \phi} \underbrace{\frac{\partial \hat{z}(s)}{\partial s}}_{=0} \nabla_z L(\hat{y}(\hat{z}), y^\star) = 0,
\end{equation}
so no gradient will flow to the encoder.
Common approaches to circumvent this problem include:
\begin{itemize}
    \item Replace the argmax node by a stochastic node where $z$ is a random
variable parametrized by $s$ (e.g., using a Gibbs distribution). Then, compute
the gradient of the \emph{expected loss} $\mathbb{E}_{z \sim p_s}[L(\hat{y}(z),
y^\star)]$. This is the approach underlying REINFORCE, score function
estimators, and minimum risk training
\citep{williams1992simple,smith2006minimum,stoyanov2011empirical}.
\citet{sparsemapcg} explore a sparse alternative to the Gibbs distribution.
    \item Keep the network deterministic, but do a continuous relaxation of the argmax node, for example replacing it with softmax or sparsemax \citep{sparsemax}. In the structured case, this gives rise to structured attention networks \citep{structured_attn} and their sparse variant, SparseMAP \citep{sparsemap}.
    Mathematically, this corresponds to moving the expectation inside the loss, optimizing $L(\hat{y}(\mathbb{E}_{z \sim p}[z]), y^\star)]$.
    \item Keep the argmax node and perform the usual forward computation, but backpropagate a surrogate gradient. This is the approach underlying straight-through estimators \citep{bengio_estimating_2013} and SPIGOT \citep{peng2018backpropagating}. We will develop this approach in the remainder of these notes.
\end{itemize}

In what follows, we assume that:
\begin{itemize}
    \item We have access to the gradient $\gamma(z) := \nabla_z L(\hat{y}(z),
y^\star)$;%
\footnote{This gradient would not exist if the decoder $g_\theta$ were defined
only for the vertices $z \in \ZZ$ and not convex combinations thereof. This
assumption is not needed in the \emph{minimum risk training} approach discussed
toward the end of this note.}
    \item We want to replace the (zero) gradient $\nabla_s L(\hat{y}(z), y^\star)$ by a surrogate $\tilde{\nabla}_s L(\hat{y}(z), y^\star)$.
\end{itemize}

\section{SPIGOT as the approximate computation of a pulled back loss}

We now provide an interpretation of SPIGOT as the minimization of a ``pulled back'' loss with respect to the latent variable $z$.
%To start gently, we first assume that $z$ is an \emph{unstructured} categorical variable,
%i.e., $\mathcal{Z} = \{e_1, \ldots, e_K\}$.
SPIGOT uses the following surrogate gradient:
\begin{eqnarray}\label{eq:spigot}
\tilde{\nabla}_s L(\hat{y}(\hat{z}), y^\star) &=&
\hat{z} - \Pi_{\conv(\ZZ)} \left[\hat{z} - \eta \nabla_z L(\hat{y}(\hat{z}), y^\star)\right]\nonumber\\
&=& \hat{z} - \mathrm{SparseMAP}(\hat{z} - \eta \gamma(\hat{z})),
\end{eqnarray}
where we used the fact that the SparseMAP transformation \citep{sparsemap} is equivalent to an
Euclidean projection, i.e.\
$\mathrm{SparseMAP}(s) = \displaystyle\arg\max_{\mu \in \conv(\ZZ)}
s^\top\mu - \frac{1}{2} \|\mu\|^2 =
\arg\min_{\mu \in \conv(\ZZ)}  \|\mu - s\|$.

%We will extend this interpretation to a structured space $\mathcal{Z}$ afterwards. In that case, the SPIGOT update takes the more general form:
%\begin{eqnarray}\label{eq:spigot_structured}
%\tilde{\nabla}_s L(\hat{y}(\hat{z}), y^\star) &=&
%\hat{z} - \Pi_{\mathrm{conv}(\mathcal{Z})} \left[\hat{z} - \eta \nabla_z L(\hat{y}(\hat{z}), y^\star)\right]\nonumber\\
%&=& \hat{z} - \textsc{SparseMAP}(\hat{z} - \eta \gamma(\hat{z})),
%\end{eqnarray}
%where $\textsc{SparseMAP}(a) = \arg\min_{\mu \in \mathrm{conv}(\mathcal{Z})} \|\mu - a\|$ is the {\bf SparseMAP transformation} \citep{sparsemap}.

\subsection{Intermediate loss on the latent variable}

Let us start by stating an obvious fact, which will draw intuition for the rest:
if we had supervision for the latent variable $z$ (e.g., if the true label $z^\star$ was revealed to us), we could simply define an {\bf intermediate loss} $\ell(\hat{z}, z^\star)$
%and minimize that directly.
which can induce nonzero updates to the encoder parameters.
In fact, if $K=|\mathcal{Z}|$ is small, we can enumerate all possible values of $z$ and define $z^\star$ as the one that minimizes the downstream loss, $z^\star = \arg\min_{z} L(\hat{y}(z), y^\star)$, using the current network parameters $\theta$; this  $z^\star$ would become our ``groundtruth.''

While this seems somewhat sensible, we may expect some instability in the beginning of the training process, since the decoder parameters $\theta$ are likely to be very suboptimal at this stage.
A more robust procedure is to allow for some label uncertainty: instead of picking a single label $z^\star \in \mathcal{Z}$,
pick
%the distribution $p \in \Delta^K$
the convex combination $\mu \in \conv(\ZZ)$
that minimizes $L(\hat{y}(\mu^\star), y^\star)$. In fact, it is likely that the
$\mu^\star$ that minimizes the downstream loss will not put all the probability
mass on a single label, and we may benefit from that if the downstream loss is what we care about.
With this in mind, we define:
\begin{equation}\label{eq:pullback_loss}
    \mu^\star = \arg\min_{\mu \in \conv(\ZZ)} L(\hat{y}(\mu), y^\star).
\end{equation}
For most interesting predictive models $\hat{y}(\mu)$, this optimization problem
is non-convex and lacks a closed form solution.
%\vlad{except for MRT in some cases, right? if L is linear?} \andre{yes, I think that's correct -- if $L$ is linear on $p$ (i.e. factored loss). But if that happens, then we're really not using $p$ as a latent variable, so it's a trivial uninteresting case}
One common strategy is the {\bf projected gradient algorithm}, which iteratively performs the following updates:
\begin{equation}\label{eq:projected_gradient}
    \mu^{(t+1)} = \Pi_{\conv(\ZZ)} \left[\mu^{(t)} - \eta_t \nabla_p
L(\hat{y}(\mu^{(t)}), y^\star)\right],
\end{equation}
where $\eta_t$ is a step size and $\Pi_{\mathcal{S}}(u) = \arg\min_{u' \in \mathcal{S}} \|u' - u\|$ denotes the Euclidean projection of point $u$ onto the set $\mathcal{S}$.
With a suitable choice of step sizes, the projected gradient algorithm is guaranteed to converge to a local optimum of Eq.~\ref{eq:pullback_loss}.
If we initialize $\mu^{(0)} = \hat{z} = \arg\max_{z \in \mathcal{Z}} s^\top z$
and run {\bf a single iteration} of projected gradient, we obtain the following
estimate $\tilde{\mu}$ of $\mu^\star$:
\begin{equation}\label{eq:single_projected_gradient}
    \tilde{\mu} = \Pi_{\conv(\ZZ)} \left[\hat{z} - \eta \nabla_p L(\hat{y}(\hat{z}), y^\star)\right].
\end{equation}
We can now treat $\tilde{\mu}$ as if it were the ``groundtruth'' label distribution, turning the optimization of the encoder $f_\phi(x)$ as if it were a {\bf supervised learning} problem.
If we use a {\bf perceptron loss},
\begin{eqnarray}\label{eq:perceptron_loss}
    \ell_{\mathrm{perc}}(\hat{z}(s), \tilde{\mu}) &=&
    \max_{z \in \mathcal{Z}} s^\top z - s^\top \tilde{\mu}\nonumber\\
    &=& s^\top \hat{z}(s) - s^\top \tilde{\mu},
\end{eqnarray}
we get the following gradient:
\begin{eqnarray}
\nabla_s \ell_{\mathrm{perc}}(\hat{z}(s), \tilde{\mu}) &=& \hat{z} - \tilde{\mu}\nonumber\\
&=& \hat{z} - \Pi_{\conv(\ZZ)} \left[\hat{z} - \eta \nabla_\mu L(\hat{y}(\hat{z}), y)\right]\nonumber\\
&=& \hat{z} - \mathrm{SparseMAP}(\hat{z} - \eta \gamma(\hat{z})),
\end{eqnarray}
which is precisely the SPIGOT gradient surrogate presented in Eq.~\ref{eq:spigot}.
This leads to the following insight into how SPIGOT updates the encoder parameters:
\begin{framed}
\begin{quote}
    SPIGOT
    minimizes the {\bf perceptron loss} between $z$ and a
    pulled back target computed by {\bf one projected gradient step}
    on $\displaystyle\min_{\mu \in \conv(\ZZ)} L(\hat{y}(\mu), y)$ starting at $\hat{z}$.
\end{quote}
\end{framed}

This construction suggests some possible alternate strategies. The first results in a well known algorithm, while the rest result in novel variations.

\begin{description}[style=unboxed,leftmargin=0cm]
    \item[Relaxing the {\boldmath $\conv(\ZZ)$} constraint.]
    The constraints in Eq.~\ref{eq:pullback_loss} make the optimization problem more
    complicated. We relax them and define
    $\mu^\star = \arg\min_{\mu \in \textcolor{purple}{\mathbb{R}^K}}
L(\hat{y}(\mu), y^\star)$.
    This problem still must be tackled iteratively, but the projection step can now
    be avoided. One iteration of gradient descent yields $\tilde{\mu} = \hat{z} -
    \eta \gamma{\hat{z}}$. The perceptron update then recovers
\textbf{straight-through},\footnote{Specifically, the ``identity'' variant
of STE, in which the backward pass acts as if $\pfrac{\hat{z}(s)}{s} =
\mathrm{Id}$ \citep{bengio_estimating_2013}.}
    via a novel derivation:
\begin{equation}
\nabla_s \ell_{\mathrm{perc}}(\hat{z}(s), \tilde{\mu}) = \hat{z} - (
\hat{z} - \eta \gamma(\hat{z})) = \eta \gamma(\hat{z}).
\end{equation}
This leads to the following insight into straight-through and its relationship
to SPIGOT:
\begin{framed}
\begin{quote}
STE minimizes the {\bf perceptron loss} between the latent $z$ and a pulled back
target computed by {\bf one gradient step} on $\displaystyle \min_{\mu \in
\mathbb{R}^K} L(\hat{y}(\mu), y)$ starting at $\hat{z}$.
\end{quote}
\end{framed}
    \item[Multiple projected gradient steps.] Instead of a single projected
gradient step, we could have run multiple steps of the iteration in
Eq.~\ref{eq:projected_gradient}. We would expect this to yield an estimate
$\tilde{\mu}$ closer to $\mu^\star$, at the cost of more computation.
    \item[Different initialization.] The projected gradient update in
Eq.~\ref{eq:single_projected_gradient} uses $\mu^{(0)} = \hat{z} = \arg\max_{z
\in \mathcal{Z}} s^\top z$ as the initial point. This is a sensible choice, if
we believe the encoder prediction $\hat{z}$ is close enough to the
``groundtruth'' $\mu^\star$, and it is computationally convenient because
$\hat{z}$ has already been computed in the forward propagation step and can be
cached. However, other initializations are possible, for example $\mu^{(0)} =
\mathrm{Marg}(s)$, or $\mu^{(0)} = 0$.
\item[Different intermediate loss function.] For simplicity, consider the
unstructured case. Let $p(s) := \mathrm{softmax}(s)$.
If we use the cross-entropy loss $\ell_{\mathrm{cross}}(p(s), \tilde{p}) = -\sum_{k=1}^K \tilde{p}_k \log p_k(s)$
instead the perceptron loss, we get
\begin{eqnarray}
\nabla_s \ell_{\mathrm{cross}}(p(s), \tilde{p}) &=& p(s) - \tilde{p}\nonumber\\
&=& p(s) - \Pi_{\Delta^K} \left[p(s) - \eta \nabla_p L(\hat{y}(p(s)), y^\star)\right]\nonumber\\
&=& p(s) - \mathrm{sparsemax}(p(s) - \eta \gamma(p(s))).
\end{eqnarray}
This generalizes easily to the CRF loss in the structured case.
    \item[Exponentiated gradient instead of projected gradient.] Also in the
unstructured case, the exponentiated gradient algorithm \citep{kivinen1997exponentiated} tackle the constrained optimization problem in Eq.~\ref{eq:pullback_loss} with the following multiplicative updates:
\begin{eqnarray}\label{eq:exponentiated_gradient}
    p^{(t+1)} &\propto& p^{(t)}\exp(-\eta_t \nabla_p L(\hat{y}(p^{(t)}), y^\star)),
\end{eqnarray}
where each point $p^{(t)}$ is strictly positive. This includes the initializer $p^{(0)}$, so we cannot have  $p^{(0)} = \hat{z}$; for this reason we assume $p^{(0)} = p(s) = \mathrm{softmax}(s)$.
A single iteration of exponentiated gradient with this initialization gives:
\begin{eqnarray}\label{eq:exponentiated_gradient_single_update}
    \tilde{p} &\propto& p(s)\exp(-\eta \nabla_p L(p(s), y^\star))\nonumber\\
    &=& \mathrm{softmax}(\log p(s) -\eta \gamma(p(s)))\nonumber\\
    &=& \mathrm{softmax}(s -\eta \gamma(p(s))).
\end{eqnarray}
With the cross-entropy loss,
i.e.\ the Kullback-Leibler divergence $\operatorname{KL}(p(s) \mid \tilde p)$,
we obtain:
\begin{eqnarray}
\nabla_s \ell_{\mathrm{cross}}(p(s), \tilde{p}) &=& p(s) - \tilde{p}\nonumber\\
&=& p(s) - \mathrm{softmax}(s -\eta \gamma(p(s)))\nonumber\\
&=& \mathrm{softmax}(s) - \mathrm{softmax}(s -\eta \gamma(p(s))),
\end{eqnarray}
i.e., the surrogate gradient is the difference of a softmax with a softmax with ``perturbed'' scores.
This generalizes to an instance of mirror descent with Kullback-Leibler
projections in the structured case.
%\vlad{should we point to the KL interpretation of this?} \andre{yes, good idea.}
\end{description}

\section{Relation to other methods for latent structure models}

\subsection{Continuous relaxation of argmax}

To simplify, let us consider the case where $z$ is a categorical variable.
If we replace the argmax node by a continuous transformation $\rho: \mathbb{R}^K \rightarrow \Delta^K$ (e.g., a softmax with a temperature), the gradient $\nabla_s L(\hat{y}(\rho(s)), y^\star)$ can be exactly computed by the chain rule:
\begin{eqnarray}
\nabla_s L(\hat{y}(\rho(s)), y) &=&
J_{\rho}(s) \nabla_z L(\hat{y}(\rho(s)), y),
\end{eqnarray}
where $J_{\rho}(s) \in \mathbb{R}^{K \times K}$ is the Jacobian of transformation $\rho$ at point $s$.

\subsection{Minimum risk training}
In this case, the network has a stochastic node $z \sim \rho(s) = p_z(s)$, with $\rho$ as above.
The gradient of the risk with respect to $s$ is:
\begin{eqnarray}
\nabla_s \mathbb{E}_s[L(\hat{y}(z), y)] &=& \sum_z L(\hat{y}(z), y) \nabla_s
p_z(s) \nonumber\\
&=& J_{\rho}(s) \ell,
\end{eqnarray}
where
$\ell \in \mathbb{R}^K$ is a vector where the $z$\textsuperscript{th} entry contains the loss value $L(\hat{y}(z), y)$.

Another way of writing the gradient above, noting that $\nabla_s p_z(s) = p_z(s)
\nabla_s \log p_z(s)$, is:
\begin{eqnarray}
\nabla_s \mathbb{E}_s[L(\hat{y}(z), y)] &=& \sum_z L(\hat{y}(z), y) \nabla_sp_z(s) \nonumber\\
&=& \mathbb{E}_s[L(\hat{y}(z), y) \nabla_s \log p_z(s)].
\end{eqnarray}
It is interesting to compare this gradient with the SPIGOT surrogate gradient in Eq.~\ref{eq:spigot}.
Also here a ``pulled-back loss'' (now $-\log p_z(s)$) is used in the gradient
computation, this time as part of a weighted sum, where the weights are the
reward $-L(\hat{y}(z), y)$ and the probability $p_z(s)$.
For example, if the downstream loss minimizer is $z^\star$ and all $z \ne z^\star$ are equally bad (i.e., if they have the same loss), then we obtain
\begin{eqnarray}
\nabla_s \mathbb{E}_s[L(\hat{y}(z), y)] &\propto& p_{z^\star}(s)  \nabla_s \log
p_{z^\star}(s).
\end{eqnarray}

\bibliographystyle{plainnat}

\end{document}